\documentclass[runningheads]{llncs}

\usepackage{eccv}

\usepackage{eccvabbrv}

\usepackage{graphicx}
\usepackage{booktabs}
\usepackage[export]{adjustbox}

\usepackage{multirow}
\usepackage{arydshln}
\usepackage{colortbl}
\usepackage[accsupp]{axessibility}  

\usepackage[pagebackref,breaklinks,colorlinks,citecolor=eccvblue]{hyperref}

\usepackage{orcidlink}

\begin{document}

\title{Monocular Gaussian SLAM with Language Extended Loop Closure } 


\author{Tian Lan\inst{1} \and
Qinwei Lin\inst{2} \and
Haoqian Wang\inst{1*}}


\institute{Tsinghua University}

\maketitle

\begin{abstract}
Recently, 3D Gaussian Splatting has shown great potential in visual Simultaneous Localization And Mapping (SLAM).
Existing methods have achieved encouraging results on RGB-D SLAM, but studies of the monocular case are still scarce.
Moreover, they also fail to correct drift errors due to the lack of loop closure and global optimization.
In this paper, we present MG-SLAM, a monocular Gaussian SLAM with a language-extended loop closure module capable of performing drift-corrected tracking and high-fidelity reconstruction while achieving a high-level understanding of the environment. 
Our key idea is to represent the global map as 3D Gaussian and use it to guide the estimation of the scene geometry, thus mitigating the efforts of missing depth information.
Further, an additional language-extended loop closure module which is based on CLIP feature is designed to continually perform global optimization to correct drift errors accumulated as the system runs.
Our system shows promising results on multiple challenging datasets in both tracking and mapping and even surpasses some existing RGB-D methods.

  \keywords{SLAM \and 3D Gaussian Splatting \and Scene Reconstruction}
\end{abstract}


\section{Introduction}
\label{sec:intro}
As a fundamental task in computer vision, visual Simultaneous Localization and Mapping (SLAM) has been extensively researched in the past few decades\cite{fuentes2015visual,cadena2016past,kazerouni2022survey}.
At present, the rapid development of applications including robotics and embodied artificial intelligence has presented new requirements apart from real-time tracking and mapping, i.e., photo-realistic scene reconstruction and a high-level understanding of the environment.

In terms of the reconstruction, early dense SLAM methods have achieved impressive dense reconstruction results with representations like surfels\cite{schops2019bad,whelan2015elasticfusion} and voxels\cite{dai2017bundlefusion,newcombe2011kinectfusion}, but most of them require RGB-D inputs and fail when depth sensors are unavailable.
In the era of deep learning, methods like\cite{zhou2018deeptam,teed2018deepv2d,teed2021droid,teed2024deep} proposed to use deep neural networks to predict the depth of pixels while overcoming the depth ambiguity and convergence, thus achieving monocular dense SLAM with depth map as representation. 

One drawback of aforementioned dense methods is the lack of ability to perform high-fidelity photo-realistic reconstruction, which is complemented by neural representation-based methods\cite{sucar2021imap,zhu2022nice,li2023dense,johari2023eslam,chung2023orbeez} after the advent of Neural Radiance Field (NeRF)\cite{mildenhall2021nerf}. 
However, such methods usually depend on a predefined bounding box, which is likely to be exceeded in runtime, \cite{matsuki2024newton,xiang2023nisb} proposed to solve this problem by constructing multiple fields but with little improvement. 
Recently proposed 3D Gaussian Splatting\cite{kerbl20233d} perfectly avoided the limitation of bounding box while achieving faster rendering and training speed.
Thus, some concurrent methods\cite{yugay2023gaussian,matsuki2023gaussian,keetha2023splatam} tried to use 3D Gaussian as their main map representation and achieved excellent results.
However, most of these methods required RGB-D inputs and suffered from the lack of a loop closure module to perform global-consistent reconstruction. 

Moreover, since the emergence of CLIP\cite{radford2021learning} broke the boundaries between images and text, \cite{kassab2023language} proposed integrating CLIP feature into a topology-based SLAM system and achieved a high-level understanding of the environment.
Inspired by that, we realized the potential application of the multi-modal model in SLAM systems and noticed the possibility of using it to perform loop detection.
It is foreseeable that methods based on such multi-modal features will have greater interpretability and a wider range of applications.  

Purposely, we present MG-SLAM, a Monocular Gaussian SLAM system with a language-extend loop closure module, that is capable of performing drift-corrected tracking and achieving high-fidelity reconstruction.
We use 3D Gaussian as the final representation of the scene and guide the estimation of the scene geometry with rendered images, thus achieving dense reconstruction results despite the lack of depth information.
Different from previous methods that perform Loop Closure (LC) with visual modal features, our proposed CLIP feature-based loop closure module can achieve a higher level of understanding of the scene and support text-to-trajectory querying while possessing full capacity for loop detection, thus correcting the drift errors.
In summary, our main contributions are as follows:
\begin{itemize}
    \item A novel CLIP feature-based loop closure module that enables a high-level understanding of the scene.
    \item A real-time mapping approach that initializes and trains 3D Gaussian representation without any additional depth information.
    \item A deep-learning-based monocular SLAM system that achieves drift-corrected tracking and high-fidelity mapping, evaluated extensively on multiple datasets ranging from real-world to synthetic, large to small scale.
\end{itemize}

\section{Related Work}
\subsubsection{Dense Visual SLAM}
Different from sparse methods\cite{mur2015orb,mur2017orb,campos2021orb,davison2007monoslam}, dense visual SLAM\cite{dai2017bundlefusion,newcombe2011kinectfusion,newcombe2011dtam,whelan2015elasticfusion,schops2019bad} focused on producing dense 3D map of the environment while achieving camera tracking.
While traditional dense methods achieved promising reconstruction results using representations such as surfels\cite{schops2019bad,whelan2015elasticfusion} and voxels\cite{dai2017bundlefusion,newcombe2011kinectfusion}, most of them relied on accurate depth information which is difficult to obtain in some situations. 
For a long time, reconstruction with RGB-only inputs remains a great challenge due to depth ambiguity and convergence.
With the advent of the deep learning era, some learning-based methods\cite{zhou2018deeptam,teed2018deepv2d,teed2021droid,teed2024deep} have made a significant breakthrough in predicting pixel depth with RGB-only inputs. 
DROID-SLAM\cite{teed2021droid}, as one of the most outstanding methods among them, proposed to predict the optical flow and corresponding confidence between frames by a recurrent neural network and use this information to perform joint optimization in a precisely designed Dense Bundle Adjustment (DBA) layer.
As the follow-up work of DROID-SLAM\cite{teed2021droid}, DPVO\cite{teed2024deep} used patches sampled from images to replace the whole frame, resulting in a more lightweight visual odometry.

However, one major deficiency of the aforementioned representations is their visualization pipelines are non-differentiable, thus preventing them from producing photo-realistic reconstruction results.
 
\subsubsection{Differentiable Rendering SLAM}
As the emergence of Neural Radiance Field\cite{mildenhall2021nerf} (NeRF),
methods such as\cite{yang2022vox,sucar2021imap,zhu2022nice,chung2023orbeez,li2023dense,zhang2023go,sandstrom2023point,rosinol2023nerf,johari2023eslam,xiang2023nisb} have achieved excellent improvement in high-fidelity reconstruction with NeRF-based representation.
As pioneering works, iMAP\cite{sucar2021imap} and Nice-SLAM\cite{zhu2022nice} implemented NeRF-based dense SLAM systems by fitting radiance fields with Multi-Layer Perceptrons (MLPs) and multi-scale feature grids, respectively.
Follow-up works like\cite{chung2023orbeez,rosinol2023nerf,li2023dense} transfer made such representation adaptable to the monocular case, and \cite{johari2023eslam} improves both efficiency and accuracy by using multi-scale feature planes.
One remaining issue of NeRF-based representation is the limitation of bounding boxes,
\cite{xiang2023nisb,matsuki2024newton} propose to dynamically construct new fields along the trajectory and smooth them in the overlapping areas while Point-SLAM\cite{sandstrom2023point} introduce point-based representation in \cite{xu2022point}.

Recently, 3D Gaussian Splatting\cite{kerbl20233d} has shown great potential in 3D reconstruction, which perfectly avoids the aforementioned limitation of NeRF-based representation and achieves faster rendering speed, thus we choose it as our primary map representation.
Note that though concurrent works\cite{keetha2023splatam,yugay2023gaussian,matsuki2023gaussian} also use 3D Gaussian representation, none of them have loop closure module in their systems and most of them require RGB-D input for good results.
This paper is completely based on monocular input and integrates a loop closure module to correct drift errors, the results are even comparable to some RGB-D-based methods.

\subsubsection{Place Recognition.}
As an essential technique for loop detection, Place Recognition can be defined as an image retrieval problem\cite{garg2021your}, i.e., given a query image, retrieving the closest match among a set of reference images.
One of the most frequently used place recognition techniques in traditional visual SLAM\cite{mur2015orb,mur2017orb,campos2021orb,galvez2012bags} is the Bag-Of-Words model, which treats each image as bags containing multiple words and determines the similarity between images by calculating the overlap of words.
The words here are also known as local descriptors of the image.
In contrast to local descriptors, global descriptors\cite{keetha2023anyloc,arandjelovic2016netvlad,hausler2021patch,jegou2010aggregating} aggregate the information of the whole image with a single vector.
VLAD\cite{jegou2010aggregating} firstly proposed the conception of aggregating extracted local descriptors as one descriptor vector and  NetVLAD\cite{arandjelovic2016netvlad} enhanced it by introducing neural networks.
Other recent follow-up works\cite{keetha2023anyloc,hausler2021patch} also improved the performance in many aspects.

In addition to methods based on visual modality, recently proposed
multi-modal\cite{radford2021learning} can also perform inter-image match while supporting cross-modal querying.
LEXIS\cite{kassab2023language} implemented a language-extended SLAM system by integrating CLIP feature, which inspired us to use such a multi-modal feature to design a loop closure module.

\section{Method}
\begin{figure}[htb]
\centering
\includegraphics[width=\textwidth]{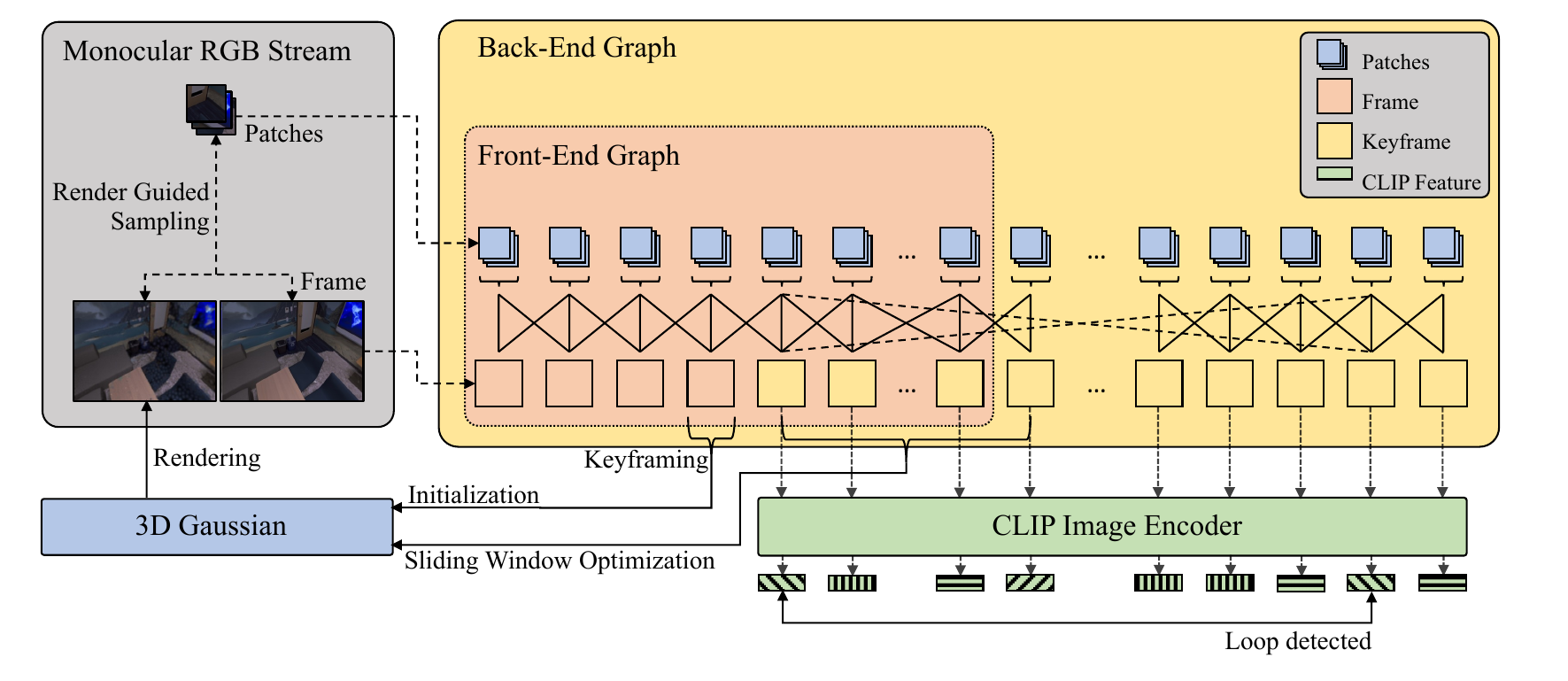}

\caption{
\textbf{System Overview.} 
Our system consists of the following components: 3D Gaussian map, CLIP feature-based loop closure module, Front-End and Back-End Graph for optimization based on DPVO\cite{teed2024deep}. 
3D Gaussian map is initialized by optimized patches and trained using keyframes within the sliding window, and the images rendered with it in turn guide the sampling of the patches.
The loop closure module continually detects loops between the current keyframe and history keyframes.
Global optimization is performed on Back-End Graph each time a new keyframe is added.
}
\label{fig:framework}

\end{figure}

As depicted in \cref{fig:framework}, our proposed MG-SLAM is extended from DPVO\cite{teed2024deep}, a deep-learning-based monocular visual odometry that uses the predicted flow of a set of sampled patches to optimize their depths as well as the camera poses jointly.
There are two major features of our system, 3D Gaussian mapping based on monocular inputs (\cref{sec:mapping}) and drift-corrected tracking via the proposed language-extended loop closure module (\cref{sec:tracking}).
We achieve the former by initializing the 3D Gaussian with optimized patches and training it with history keyframes within a sliding window.
Additionally, we use the image rendered with 3D Gaussian to guide the sampling of new patches, thus improving the reconstruction quality.
As for the latter, inspired by \cite{kassab2023language}, we design a language-extended loop closure module based on CLIP\cite{radford2021learning} features which is not only capable of loop detection but also supports a high level of understanding of the environment.
Combined with this loop closure module, we continually build a Back-End Graph and perform global optimization over it, thus enabling drift-corrected tracking.

Formally, our algorithm takes monocular RGB sequence $\{I_{i}\}_{i=1}^{N}$ as input, and outputs drift-corrected trajectories $\{\xi_{i}\}_{i=1}^{N}$ and global map represented by 3D Gaussian $\{G\}$.

\subsection{Preliminary}
\subsubsection{Patch-based Visual Odometry.}
 Similar to DROID-SLAM\cite{teed2021droid}, DPVO\cite{teed2024deep} consists of two main steps: Graph Construction and Network Update.
The graph it maintained is a bipartite patch-frame graph which consists of frame node $\{I_{j}\}_{j=1}^{N}$, patch node $\{P_{k}=(x_{k},y_{k},d_{k})\}_{k=1}^{M}$ and edges $\mathcal{E}$ connect them together. 
Here, $d_{k}$ denotes the inverse depth of $P_{k}$, and $(x_{k},y_{k})$ means the pixel coordinates on the frame $I_i$ where it is sampled. 
During inference, the entire graph will be fed into the network to predict the confidence weights $w_{kj}$ and residuals $r_{kj}$ corresponding to each edge $(k,j) \in \mathcal{E}$, after that the weights and residuals will be passed to the DBA (Dense Bundle Adjustment) layer to optimize the pose $\xi_j$ of each frame $I_j$ and the inverse depth $d_k$ of each patch $P_k$ by minimizing following objective:
\begin{equation}
  \sum_{(k,j) \in \mathcal{E}}{\|p_{kj}-(\hat{p}_{kj}+r_{kj}) \|}_{w_{kj}}^{2},
\label{eq:dba}
\end{equation}
where $p_{kj}$ is the pixel coordinates obtained by reprojecting patch $P_{k}$ from frame $I_{i}$ to frame $I_{j}$ and $\hat{p}_{kj}$ is the center coordinates of $p_{kj}$.

\subsubsection{3D Gaussian Representation.}
We use 3D Gaussian as the final representation of the reconstructed scene. Original 3D Gaussian\cite{kerbl20233d} contains positions $\mu \in \mathbf{R}^{3}$, rotation $q \in \mathbf{R}^{4}$, scale $s \in \mathbf{R}^{3} $, opacity $o \in \mathbf{R}$ and SH coefficient $\mathbf{SH} \in \mathbf{R}^{48}$, in concerns of memory consumption and optimization speed, we choose to optimize the color of 3D Gaussian $c \in \mathbf{R}^{3}$ directly instead of SH coefficient $ \mathbf{SH} $. During runtime, we initialize aforementioned 3D Gaussian representations $\{G_k=(\mu_k,q_k,s_k,o_k,c_k)\}_{k=1}^{K}$ with optimized patches' depth $\{d_j\}_{j=1}^{M}$ and use estimated poses $\{\xi_{i}\}_{i=1}^{N}$ to train them.

\subsection{3D Gaussian Mapping}
\label{sec:mapping}
The main representation of our proposed MG-SLAM is 3D Gaussian, which is capable of photo-realistic reconstruction.
We first initialize new Gaussians with optimized patches, then a sliding window optimization strategy is introduced to mitigate the influence of the sparsity of training views.
Further, to achieve a higher quality of reconstruction with finite Gaussians, we design a render-guided sampling approach that uses rendered images to adaptively choose the position where new patches are sampled.

\subsubsection{Gaussian Initialization.}
For Gaussian initialization, after a new frame is added and optimized for 4 iterations, we initialize new Gaussians with patches sampled from it. 
Given patches $\{P=(x,y,d)\}$ sampled from frame $I_i$, firstly we back-project the center of each patch $(\hat{x},\hat{y},\hat{d})$ to obtain Gaussian's position $\mu$:
\begin{equation}
\mu=R_{i}\begin{bmatrix}
\frac{\hat{x}-c_{x}}{\hat{d}f_{x}} & 
\frac{\hat{x}-c_{x}}{\hat{d}f_{x}} &
\frac{1}{\hat{d}}
\end{bmatrix}^{T}+t_i
\label{eq:backproj}
\end{equation}
where $f_x,f_y,c_x,c_y$ is the intrinsic of camera and $[R_i,t_i]$ denotes the pose of $I_i$.
Then $\hat{d}$ will be used to initialize Gaussian's scales so that its projection on frame $I_i$ is one pixel in size:
\begin{equation}
s=\begin{bmatrix}
    \frac{\hat{d}}{f_x} & \frac{\hat{d}}{f_y} & \frac{2\hat{d}}{f_x+f_y}
\end{bmatrix}^{T}
\label{eq:scale}
\end{equation}
Finally, the opacity of Gaussian will be initialized as 0.5 and the color of Gaussian $c$ will be initialized as the color of patch center $(\hat{x},\hat{y})$ on frame $I_i$. 

\subsubsection{Render-Guided Sampling.}
According to DPVO\cite{teed2024deep}, randomly selecting patch centroids is superior to using regular 2D key points as patch centroids in terms of tracking accuracy. 
However, randomly picking patches gives the highest performance in tracking, Gaussians initialized by these patches are usually distributed very unevenly, thus making the mapping process unstable.
To make the sampling of patches more suitable with 3D Gaussian mapping, we make a compromise on the tracking performance and propose our rendering-guided sampling technique, which could adaptively make the initialized Gaussians evenly distributed in 3D space.

Briefly, after 3D Gaussian map is initialized, we first use the estimated pose $\hat{\xi}_{i}$ of the latest frame $I_i$ to render the image $\hat{I}_i$.
Then the per-pixel RGB L1 loss will be calculated between $\hat{I}_i$ and ground truth image $I_i$.
After that, the pixel with the highest loss will be chosen as the centroid of the newly sampled patch and its neighboring pixels will be suppressed to avoid too close proximity between sampled patches.
We repeat the process above until enough patches have been sampled.

In practice, in order to prevent patches from clustering in unknown regions, we split the image into multiple grids and perform render-guided sampling separately within each grid.

\subsubsection{Sliding Window Optimization.}
To ensure the multi-view consistency of 3D Gaussian and prevent it from overfitting, we adopt a sliding window optimization strategy for training.
That is, each time a new keyframe $I_i$ is identified, we optimize 3D Gaussian using $I_i$ and $n$ adjacent keyframes $\{I_j\}_{j=i-n}^{i-1}$ as supervision.
Following \cite{kerbl20233d}, we use a combination of RGB L1 loss and SSIM loss as the primary loss:
\begin{equation}
\mathcal{L}_{color}=\sum_{I_i \in \mathcal{W}}{(1-\lambda)\|I_i-\hat{I}_i\|_{1}+\lambda(1-\text{SSIM}(I_i,\hat{I}_i))}
\label{eq:closs}
\end{equation}
where $\mathcal{W}$ denotes the sliding window containing adjacent keyframes and $\hat{I}_i$ means the rendered image corresponding to $I_i$. 
$\lambda=0.2$ is the ratio between L1 loss and SSIM loss.
Since we only use keyframes for training, to prevent artifacts resulting from sparse views, we introduce an additional L2 regular loss $\mathcal{L}_{reg}$ to penalize scales larger than threshold $\tau_{s}$.

We use $\mathcal{L}=\mathcal{L}_{color}+\lambda_{reg}\mathcal{L}_{reg}$ as the final loss function to optimize Gaussian, where $\lambda_{reg}$ is set to 1. Similar to \cite{kerbl20233d}, we prune Gaussians with opacity less than 0.005, but the difference is that we do not perform any densification on Gaussians.
 
In the end, since we only optimize maps within the sliding window, previous maps may suffer from oblivion during the process.
To avoid this problem, after the tracking process completes, we perform several iterations of optimization using all the keyframes.

\subsection{Drift-corrected Tracking}
\label{sec:tracking}
As mentioned earlier, we build our SLAM system on top of DPVO\cite{teed2024deep}, which is a remarkable lightweight deep learning-based visual odometry work. 
However, despite its advantages of high accuracy and speed, as a visual odometry, it inevitably suffers from drift errors that degrade tracking performance over long sequences.
Purposely, we propose a loop closure module based on the CLIP\cite{radford2021learning} feature that detects loops in the trajectory while supporting a high-level understanding of the environment, i.e., text-to-trajectory querying.
Further, to make global optimization feasible under the framework of DPVO\cite{teed2024deep}, an additional divide-and-conquer scheme is proposed to alleviate the memory consumption caused by the massive graph structure.

\subsubsection{Loop Detection.}
The goal of Loop Detection is to check whether the trajectory passes through the same location repeatedly, which is analogous to Place Recognition.
Traditional VSLAM methods\cite{mur2015orb,mur2017orb,campos2021orb} usually use Bag-Of-Words models\cite{galvez2012bags} to model the probability that different frames belong to the same place, while learning-based Place Recognition methods tend to use deep neural networks. 
We choose CLIP\cite{radford2021learning} to extract features of frames for Place Recognition since it not only describes similarities between images but also between text and images, thus supporting a high-level understanding of trajectory.  

For each new keyframe $I$, we first extract its feature vector $f$ using CLIP\cite{radford2021learning} image encoder. 
By calculating cosine similarities between the current keyframe and all history keyframes, keyframes with similarities greater than threshold $\tau_{sim}$ will be recognized as loop candidates. 
Finally, we compute the optical flow magnitude between the current keyframe and loop candidates, discarding incorrectly recognized candidates whose optical flow is larger than threshold $\tau_{flow}$ and keeping the remaining candidates as detected loop frames.

\subsubsection{Global Graph Construction.} 
In order to perform global optimization, we maintain a new patch-frame graph in addition to DPVO\cite{teed2024deep}'s graph.
For the convenience of discussion, here we call the graph maintained by DPVO\cite{teed2024deep} the Front-End Graph, and the new graph we maintained the Back-End Graph.
Back-End Graph and Front-End Graph are consistent in adding new frames and deleting non-keyframes, the difference is that the Back-End Graph will not delete the edges that exceed the removal window and the loop edges will only be added to the Back-End Graph.

For loop edges addition, given a new keyframe $I_{i}$ and detected loop frames $\{I_{j}\}_{j < i-R_{recent}}$ where $R_{recent}$ is the radius to suppress frames too close to $I_{i}$, we connect patches sampled from $I_i$ to $I_j$ and connect patches sampled from $I_j$ to $I_i$ conversely.

\subsubsection{Back-End Optimization.}
One challenge to performing global optimization is that the scale of the Back-End Graph increases as the input sequence grows, and embedding the whole graph into the network will lead to memory overflow.
To address this problem, we additionally record the frame index for each patch it was sampled from, and then we partition the edges in the Back-End Graph into multiple subsets according to the frame index of connected patches.
These subsets will be fed into the network in batches to get corresponding weights and residuals, which will be spliced together to perform global optimization in DBA layers.
\begin{figure}[h]
\centering
\includegraphics[width=\textwidth]{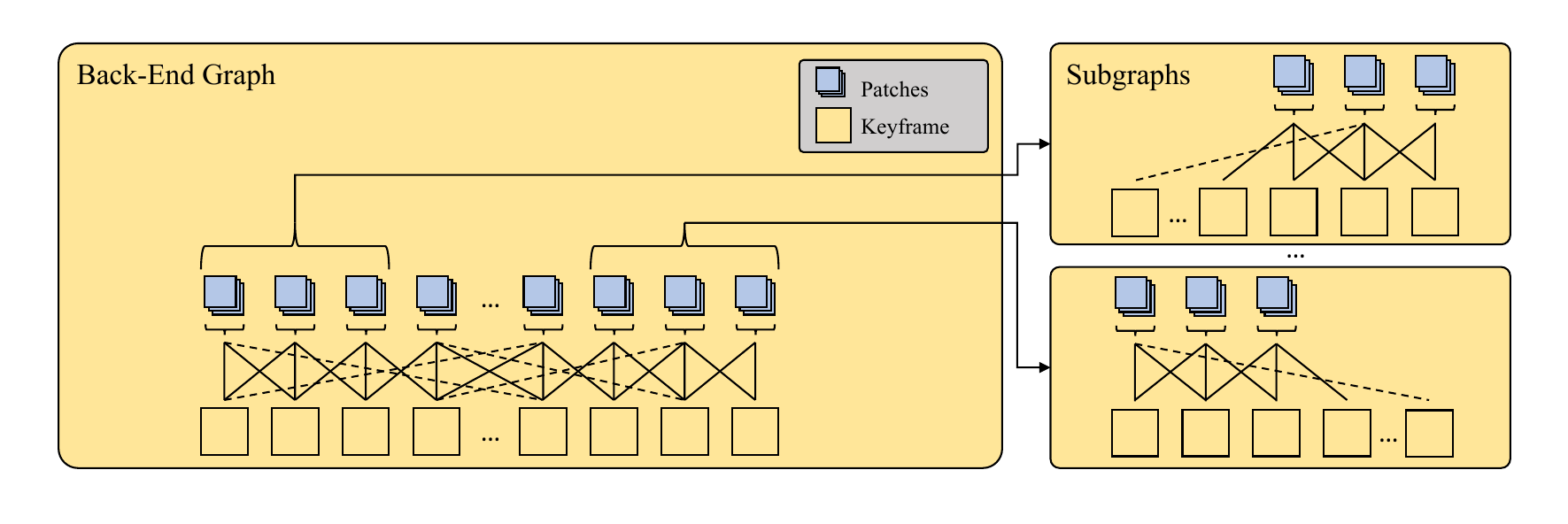}
\caption{
\textbf{Illustration of Subgraph Partition.}
The edges are grouped by the index of the frame where its connected patch is sampled from. 
Each group of edges and their connected nodes form a subgraph.
}
\label{fig:partition}
\end{figure}
While such processing might have subtle differences compared with the original implementation in DPVO\cite{teed2024deep}, it saves lots of memory and doesn't consume too much time. 
In our system, loop detection will be triggered whenever a new keyframe is added, along with the global optimization based on the Back-End Graph.

\section{Experiment}
In this section, we evaluate our system quantitatively on multiple datasets of various scales, both real and synthetic.
Qualitative results and ablation experiments are also presented.
\subsection{Experimental Setup}
\subsubsection{Datasets.}
We test our system on several diverse datasets including Replica\cite{straub2019replica}, ScanNet\cite{dai2017scannet}, TUM RGB-D\cite{sturm2012benchmark} and EuRoC\cite{burri2016euroc}.
The Replica\cite{straub2019replica} dataset is a synthetic dataset that provides high-fidelity 3D models of several different indoor scenes, here we use the same rendered RGB sequence as in \cite{zhu2022nice}. 
The ScanNet\cite{dai2017scannet} dataset is a real-world dataset containing thousands of RGB-D sequences with different lengths and we test our method on several representative sequences. 
The TUM RGB-D\cite{sturm2012benchmark} benchmark is a small-scale indoor dataset containing RGB-D sequences recorded by the Microsoft Kinect sensor along the ground truth trajectory.
The EuRoC\cite{burri2016euroc} dataset consists of 11 indoor stereo sequences collected by a micro aerial vehicle (MAV).

\subsubsection{Baselines.}
We compare with concurrent Gaussian SLAM methods\cite{keetha2023splatam,yugay2023gaussian,matsuki2023gaussian}, where \cite{keetha2023splatam,yugay2023gaussian} are based on RGBD input only and \cite{matsuki2023gaussian} additionally support monocular input.
As the only work that is currently open-source, we compare most of our results with SplaTAM\cite{keetha2023splatam} even though it is based on RGBD input.
Notably, Gaussian Splatting SLAM\cite{matsuki2023gaussian} supports monocular input but since the code is not released yet, we can only use part of the result from the paper for comparison.
We also make a comparison with GO-SLAM\cite{zhang2023go} cause it also includes global optimization and loop closure module.
In addition, some previous NeRF-based methods\cite{zhu2022nice,sandstrom2023point,yang2022vox} are also involved in the comparison.
For fair comparisons,
all the results are provided by the authors directly or generated by the released code under
the default settings and take the median of 5 trials.

\subsubsection{Metrics.}
Following\cite{teed2021droid,zhang2023go,mur2017orb}, we first align the estimated trajectory with ground truth and then evaluate the camera pose accuracy by Absolute Trajectory Error (ATE) RMSE.
Metrics including Peak Signal-to-Noise Ratio (PSNR), Structural Similarity Index (SSIM), and Learned Perceptual Image Patch Similarity (LPIPS) are used to measure the rendering quality of reconstruction results.
If not specified, we report the median results of 5 runs following \cite{teed2024deep}.

\subsubsection{Implementation Details.}
We test our system on a PC with a 3.6 GHz Intel Core i7-12700K CPU and an NVIDIA RTX 4090 GPU. 
We use pre-trained weights provided by DPVO\cite{teed2024deep} in the tracking network and follow DPVO-Default configuration, sampling 96 patches per frame and tracking in 10 frames optimization frames.
The length of sliding windows in 3D Gaussian mapping is set to 7, all the input images are resized to 640 $\times$ 480.
The learning rates of Gaussian are consistent with \cite{keetha2023splatam}.

\begin{figure}[h]
\centering
\includegraphics[width=\textwidth]{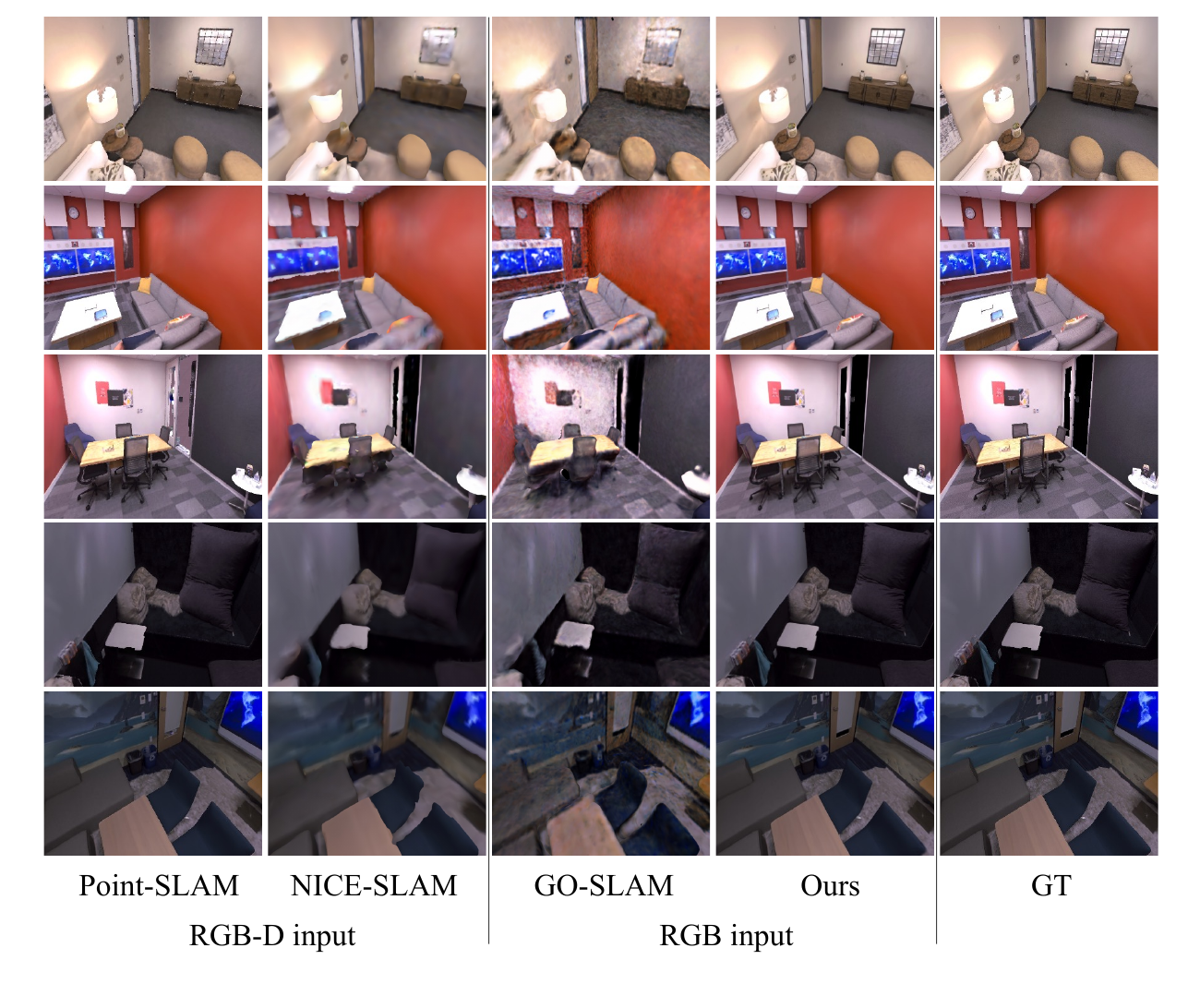}
\caption{
\textbf{Rendering Performance on Replica\cite{straub2019replica}.}
Thanks to 3D Gaussian representation, our method outperforms previous NeRF SLAM methods based on both RGB-D\cite{xu2022point,zhu2022nice} and RGB\cite{zhang2023go}. }
\label{fig:render_result}
\end{figure}
\begin{figure}[h]
\centering
\includegraphics[width=\textwidth]{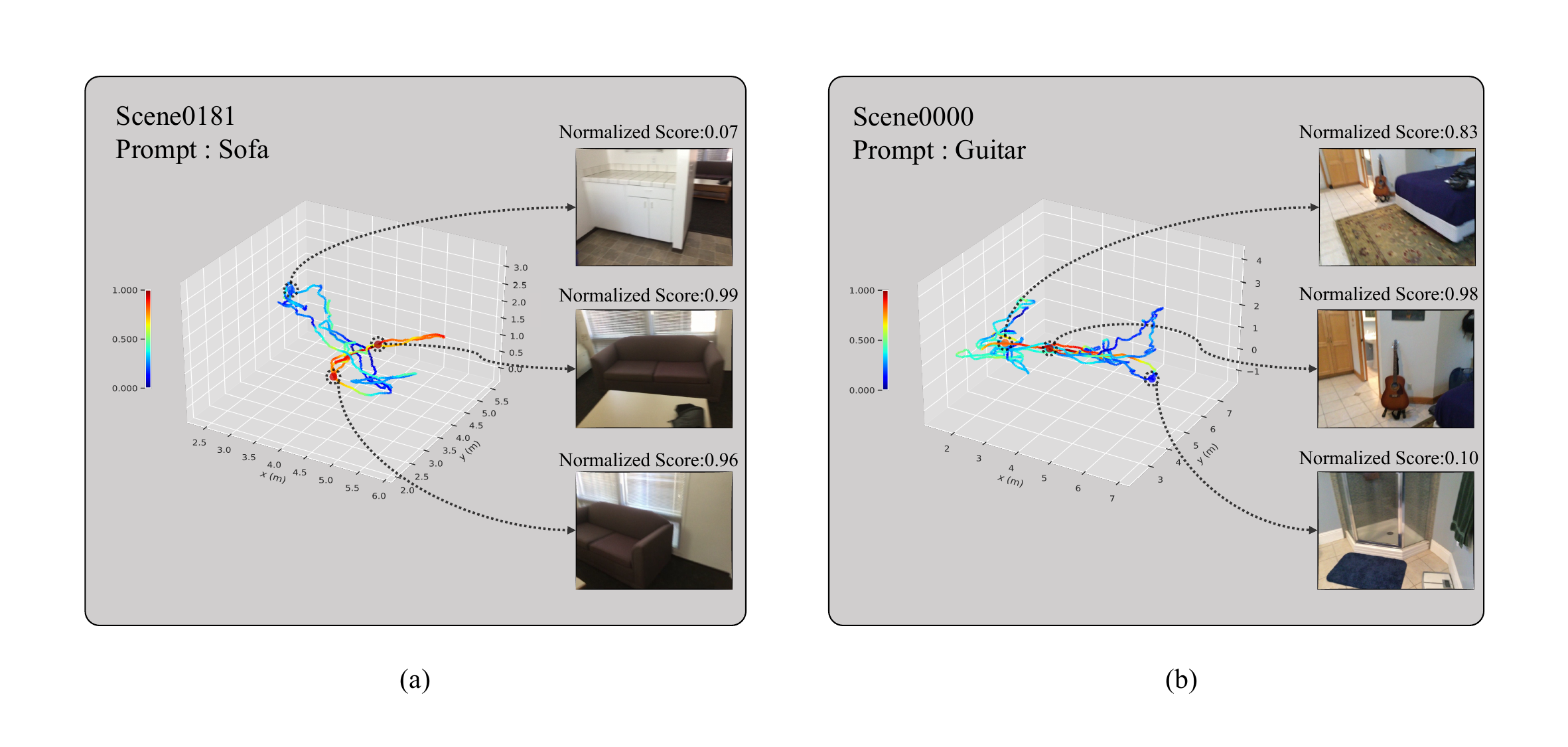}
\caption{
\textbf{Example on Text-to-Trajectory Querying.}
Given a text prompt, our system is able to return the most relevant keyframe along the trajectory.}
\label{fig:query_result}
\end{figure}
\begin{table*}[htb]   
\begin{center}
\small
 \caption{
 \textbf{ATE RMSE[m] on Replica\cite{straub2019replica} dataset. }
 Results of \cite{yang2022vox,zhu2022nice,xu2022point,keetha2023splatam} are taken from \cite{keetha2023splatam} and results of \cite{teed2021droid} and \cite{zhang2023go} are obtained by running their codes.
 Cell color indicates \colorbox{SpringGreen!50}{best}, \colorbox{Yellow!50}{second best} and \colorbox{Apricot!50}{third best}.
 }
 \label{tab:replica_ate}
\renewcommand\arraystretch{1.2} 
\setlength{\tabcolsep}{1.0mm}{ 
\resizebox{\textwidth}{!}{
\begin{tabular}{c c ccccccccc}\toprule
\multicolumn{2}{c}{Method} & Office0 & Office01 & Office02 & Office03 & Office04 & Room0 & Room1 & Room2 & Avg\\ \midrule
\multirow{4}{*}{\rotatebox{90}{RGB-D}} 
&Vox-Fusion\cite{yang2022vox}   
&0.0848&0.0204&0.0258&0.0111&0.0294&0.0137&0.0470&0.0147&0.0309 \\
&Nice-SLAM\cite{zhu2022nice}        
&0.0088&0.0100&0.0106&0.0110&0.0113&0.0097&0.0131&0.0107&0.0106   \\ 
&Point-SLAM\cite{sandstrom2023point}        
&\cellcolor{Apricot!50}{0.0038}&0.0048&0.0054&0.0069&\cellcolor{Apricot!50}{0.0072}&0.0061&\cellcolor{Apricot!50}{0.0041}&0.0037&0.0052   \\
&SplaTAM\cite{keetha2023splatam}    
&0.0047&\cellcolor{Yellow!50}{0.0027}&\cellcolor{SpringGreen!50}{0.0029}&\cellcolor{SpringGreen!50}{0.0032}&\cellcolor{Yellow!50}{0.0055}&\cellcolor{SpringGreen!50}{0.0031}&\cellcolor{Yellow!50}{0.0040}&\cellcolor{SpringGreen!50}{0.0029}&\cellcolor{Yellow!50}{0.0036}     \\
\midrule

\multirow{3}{*}{\rotatebox{90}{Mono.}}  
&DROID-SLAM\cite{teed2021droid}
&0.0075 
&0.0077         
&0.0102     
&0.0101     
&0.0107     
&0.0059     
&0.0068    
&0.0048    
&0.0079    \\ 
&GO-SLAM\cite{zhang2023go}
&\cellcolor{Yellow!50}{0.0032}&\cellcolor{Apricot!50}{0.0044}&\cellcolor{Yellow!50}{0.0038}&\cellcolor{Apricot!50}{0.0058}&0.0073&\cellcolor{Apricot!50}{0.0044}&0.0047&\cellcolor{Apricot!50}{0.0032}&\cellcolor{Apricot!50}{0.0046}   \\
&Ours        
&\cellcolor{SpringGreen!50}{0.0029}
&\cellcolor{SpringGreen!50}{0.0023}
&\cellcolor{Apricot!50}{0.0043}
&\cellcolor{Yellow!50}{0.0043}
&\cellcolor{SpringGreen!50}{0.0050}&\cellcolor{Yellow!50}{0.0040}&\cellcolor{SpringGreen!50}{0.0037}&\cellcolor{Yellow!50}{0.0031}&\cellcolor{SpringGreen!50}{0.0032}  \\ \bottomrule
\end{tabular}}}
\end{center}
\end{table*}

\begin{table*}[htb]
\begin{center}
\small
 \caption{
 \textbf{Rendering Metrics on Replica dataset.}
 Results of \cite{yang2022vox,zhu2022nice,sandstrom2023point,keetha2023splatam} are taken from \cite{keetha2023splatam}.
 Cell color indicates \colorbox{SpringGreen!50}{best}, \colorbox{Yellow!50}{second best} and \colorbox{Apricot!50}{third best}.}
 \label{tab:replica_render}
\renewcommand\arraystretch{1.2} 
\setlength{\tabcolsep}{1.0mm}{ 
\resizebox{\textwidth}{!}{
\begin{tabular}{c c c ccccc ccc}\toprule
Method &Metric  & Office0 & Office01 & Office02 & Office03 & Office04 & Room0 & Room1 & Room2 &Avg \\ \midrule

\multirow{3}{*}{Vox-Fusion\cite{yang2022vox}} 
&$\text{PSNR}\uparrow$ 
&27.79&29.83&20.33&23.47&25.21&22.39&22.36&23.92&24.41   \\
&$\text{SSIM}\uparrow$    
&0.86&0.88&0.79&0.80&0.85&0.68&0.75&0.80&0.80    \\
&$\text{LPIPS}\downarrow$        
&0.24&\cellcolor{Apricot!50}{0.18}&0.24&0.21&\cellcolor{Apricot!50}{0.20}&0.30&0.27&0.23&0.24  \\   \midrule

\multirow{3}{*}{NICE-SLAM\cite{zhu2022nice}} 
&$\text{PSNR}\uparrow$ 
&29.07&30.34&19.66&22.23&24.94&22.12&22.47&24.52&24.42   \\
&$\text{SSIM}\uparrow$    
&0.87&0.89&0.80&0.80&0.86&0.69&0.76&0.81&0.81    \\
&$\text{LPIPS}\downarrow$        
&0.23&\cellcolor{Apricot!50}{0.18}&0.24&0.21&\cellcolor{Apricot!50}{0.20}&0.33&0.27&\cellcolor{Apricot!50}{0.21}&0.23  \\   \midrule

\multirow{3}{*}{Point-SLAM\cite{sandstrom2023point}}  
&$\text{PSNR}\uparrow$        
&\cellcolor{Yellow!50}{38.26}&\cellcolor{Yellow!50}{39.16}&\cellcolor{Yellow!50}{33.99}&\cellcolor{SpringGreen!50}{33.48}&\cellcolor{SpringGreen!50}{33.49}&\cellcolor{Yellow!50}{32.40}&\cellcolor{SpringGreen!50}{34.08}&\cellcolor{SpringGreen!50}{35.50}&\cellcolor{SpringGreen!50}{35.17}  \\
&$\text{SSIM}\uparrow$        
&\cellcolor{SpringGreen!50}{0.98}&\cellcolor{SpringGreen!50}{0.99}&\cellcolor{Yellow!50}{0.96}&\cellcolor{SpringGreen!50}{0.96}&\cellcolor{SpringGreen!50}{0.98}&\cellcolor{Yellow!50}{0.97}&\cellcolor{SpringGreen!50}{0.98}&\cellcolor{SpringGreen!50}{0.98}&\cellcolor{SpringGreen!50}{0.98}    \\ 
&$\text{LPIPS}\downarrow$        
&\cellcolor{Yellow!50}{0.10}&\cellcolor{Yellow!50}{0.12}&\cellcolor{Yellow!50}{0.16}&\cellcolor{Yellow!50}{0.13}&\cellcolor{SpringGreen!50}{0.14}&\cellcolor{Yellow!50}{0.11}&\cellcolor{Yellow!50}{0.12}&\cellcolor{Yellow!50}{0.11}&\cellcolor{Yellow!50}{0.12}  \\   \midrule

\multirow{3}{*}{SplaTAM\cite{keetha2023splatam}}  
&$\text{PSNR}\uparrow$        
&\cellcolor{Yellow!50}{38.26}&\cellcolor{SpringGreen!50}{39.17}&\cellcolor{SpringGreen!50}{39.17}&\cellcolor{Apricot!50}{29.70}&\cellcolor{Yellow!50}{31.81}&\cellcolor{SpringGreen!50}{32.86}&\cellcolor{Yellow!50}{33.89}&\cellcolor{Yellow!50}{35.25}&\cellcolor{Yellow!50}{34.11}  \\
&$\text{SSIM}\uparrow$        
&\cellcolor{SpringGreen!50}{0.98}&\cellcolor{Yellow!50}{0.98}&\cellcolor{SpringGreen!50}{0.97}&\cellcolor{Yellow!50}{0.95}&\cellcolor{Yellow!50}{0.95}&\cellcolor{SpringGreen!50}{0.98}&\cellcolor{Yellow!50}{0.97}&\cellcolor{SpringGreen!50}{0.98}&\cellcolor{Yellow!50}{0.97}\\ 
&$\text{LPIPS}\downarrow$        
&\cellcolor{SpringGreen!50}{0.09}&\cellcolor{SpringGreen!50}{0.09}&\cellcolor{SpringGreen!50}{0.10}&\cellcolor{SpringGreen!50}{0.12}&\cellcolor{Yellow!50}{0.15}&\cellcolor{SpringGreen!50}{0.07}&\cellcolor{SpringGreen!50}{0.10}&\cellcolor{SpringGreen!50}{0.08}&\cellcolor{SpringGreen!50}{0.10} \\   \midrule

\multirow{3}{*}{Ours}  
&$\text{PSNR}\uparrow$        
&\cellcolor{SpringGreen!50}{38.57}&\cellcolor{Apricot!50}{38.87}&\cellcolor{Apricot!50}{32.21}&\cellcolor{Yellow!50}{31.65}&\cellcolor{Yellow!50}{31.81}&\cellcolor{Apricot!50}{30.35}&\cellcolor{Apricot!50}{31.65}&\cellcolor{Apricot!50}{33.15}&\cellcolor{Apricot!50}{33.59}   \\ 
&$\text{SSIM}\uparrow$       
&\cellcolor{Apricot!50}{0.96}&\cellcolor{Apricot!50}{0.96}&\cellcolor{Apricot!50}{0.93}&\cellcolor{Apricot!50}{0.93}&\cellcolor{Apricot!50}{0.94}&\cellcolor{Apricot!50}{0.88}&\cellcolor{Apricot!50}{0.91}&\cellcolor{Apricot!50}{0.94}&\cellcolor{Apricot!50}{0.93}   \\ 
&$\text{LPIPS}\downarrow$        
&\cellcolor{Apricot!50}{0.16}&0.21&\cellcolor{Apricot!50}{0.23}&\cellcolor{Apricot!50}{0.22}&0.21&\cellcolor{Apricot!50}{0.23}&\cellcolor{Apricot!50}{0.26}&0.23&\cellcolor{Apricot!50}{0.22}   \\ \bottomrule
\end{tabular}}}
\end{center}
\end{table*}

\begin{table*}[htb]
\begin{center}
\small
 \caption{
 \textbf{ATE RMSE[m] on ScanNet dataset.}
 Results of \cite{yang2022vox,zhu2022nice,sandstrom2023point,keetha2023splatam} are taken from \cite{keetha2023splatam} and results of \cite{teed2021droid,zhang2023go} are taken from \cite{zhang2023go}.
 Cell color indicates \colorbox{SpringGreen!50}{best}, \colorbox{Yellow!50}{second best} and \colorbox{Apricot!50}{third best}.}
 \label{tab:scannet_ate}
\renewcommand\arraystretch{1.2} 
\setlength{\tabcolsep}{1.0mm}{ 
\resizebox{8cm}{!}{    
\begin{tabular}{c c ccccccccc}\toprule
\multicolumn{2}{c}{Method}  & 0000  & 0059 & 0106 & 0169 & 0181 & Avg \\ \midrule
\multirow{4}{*}{\rotatebox{90}{RGB-D}} 
&Vox-Fusion\cite{yang2022vox}   
& 0.6884&0.2418&0.0841&0.2728&0.2330&0.3040    \\
&Nice-SLAM\cite{zhu2022nice}        
&0.1200&0.1400&0.0790&0.1090&0.1340&0.1164  \\ 
&Point-SLAM\cite{sandstrom2023point}        
&0.1024&\cellcolor{SpringGreen!50}{0.0781}&0.0865&0.2216&0.1477&0.1272   \\
&SplaTAM\cite{keetha2023splatam}    
&0.1283&0.1010&0.1772&0.1208&0.1110&0.1276 \\
\midrule

\multirow{3}{*}{\rotatebox{90}{Mono.}}  
&DROID-SLAM\cite{teed2021droid}
&\cellcolor{SpringGreen!50}{0.0548}&0.0900&\cellcolor{SpringGreen!50}{0.0676}&\cellcolor{SpringGreen!50}{0.0786}&\cellcolor{SpringGreen!50}{0.0741}&\cellcolor{SpringGreen!50}{0.0730}        \\ 
&GO-SLAM\cite{zhang2023go}
&\cellcolor{Apricot!50}{0.0594}&\cellcolor{Yellow!50}{0.0827}&\cellcolor{Apricot!50}{0.0807}&\cellcolor{Yellow!50}{0.0842}&\cellcolor{Apricot!50}{0.0829}&\cellcolor{Apricot!50}{0.0779}    \\
&Ours        
&\cellcolor{Yellow!50}{0.0552}&\cellcolor{Apricot!50}{0.0845}&\cellcolor{Yellow!50}{0.0678}&\cellcolor{Apricot!50}{0.0851}&\cellcolor{Yellow!50}{0.0746}&\cellcolor{Yellow!50}{0.0734}   \\ \bottomrule
\end{tabular}}}
\end{center}
\end{table*}

\begin{table}[htb]
\begin{center}
\small
 \caption{
 \textbf{ATE RMSE[m] on TUM RGB-D benchmark.} 
 Results of \cite{whelan2015elasticfusion,mur2017orb,yang2022vox,zhu2022nice,sandstrom2023point,keetha2023splatam} are taken from \cite{keetha2023splatam}.
 Results of \cite{matsuki2023gaussian} are partially blank since they are not provided in the paper.
 Cell color indicates \colorbox{SpringGreen!50}{best}, \colorbox{Yellow!50}{second best} and \colorbox{Apricot!50}{third best}.}
 \label{tab:tum_ate}
\renewcommand\arraystretch{1.2} 
\setlength{\tabcolsep}{1.0mm}{ 
\resizebox{10cm}{!}{    
\begin{tabular}{c c cccccc}\toprule
\multicolumn{2}{c}{Method} & fr1/desk & fr1/desk2 & fr1/room & fr2/xyz & fr3/off & Avg  \\ \midrule
\multirow{6}{*}{\rotatebox{90}{RGB-D}}

&ElasticFusion\cite{whelan2015elasticfusion}  
&0.0253&0.0683&0.2149&0.0117&0.0252&0.0691    \\

&ORBSLAM2\cite{mur2017orb}  
&\cellcolor{Yellow!50}{0.0160}&\cellcolor{SpringGreen!50}{0.0220}&\cellcolor{Yellow!50}{0.0470}&\cellcolor{SpringGreen!50}{0.0040}&\cellcolor{SpringGreen!50}{0.0100}&\cellcolor{SpringGreen!50}{0.0198} \\

&Vox-Fusion\cite{yang2022vox}  
&0.0352&0.0600&0.1953&0.0149&0.2601&0.1131       \\

&Nice-SLAM\cite{zhu2022nice}        
&0.0426&0.0499&0.3449&0.3173&0.0387&0.1587       \\

&Point-SLAM\cite{sandstrom2023point}        
&0.0434&0.0454&0.3092&0.0131&0.0348&0.0892       \\

&SplaTAM\cite{keetha2023splatam}    
&0.0335&0.0654&\cellcolor{Apricot!50}{0.1113}&0.0124&0.0516&0.0548        \\
 \midrule
\multirow{4}{*}{\rotatebox{90}{Mono.}}  

&DROID-SLAM\cite{teed2021droid}        
&\cellcolor{Apricot!50}{0.0177}
&\cellcolor{Yellow!50}{0.0267}
&\cellcolor{SpringGreen!50}{0.0433}
&\cellcolor{Apricot!50}{0.0046}
&\cellcolor{Apricot!50}{0.0289}
&\cellcolor{Yellow!50}{0.0242}      \\

&GO-SLAM\cite{zhang2023go}        
&\cellcolor{SpringGreen!50}{0.0150} 
&0.0487
&0.4096
&\cellcolor{Yellow!50}{0.0042}
&\cellcolor{Yellow!50}{0.0210}
&0.0997      \\

&GaussianSplatting-SLAM\cite{matsuki2023gaussian}        
&0.0415 &   -  &  -   &0.0479      &0.0439      &-        \\ 

&Ours        
&0.0282&\cellcolor{Apricot!50}{0.0357}&0.1641&0.0082&0.0322&\cellcolor{Apricot!50}{0.0537}       \\ 

\bottomrule
\end{tabular}}}
\end{center}
\end{table}

\begin{table*}[htb]
\begin{center}
\small
 \caption{
 \textbf{ATE[m] on EuRoC dataset.}
 Results of \cite{mur2015orb,mur2017orb,campos2021orb,teed2021droid,zhang2023go} are taken from \cite{zhang2023go}.
 Results of \cite{mur2015orb,mur2017orb,campos2021orb} are partially blank because they fail in these scenarios. 
  Cell color indicates \colorbox{SpringGreen!50}{best}, \colorbox{Yellow!50}{second best} and \colorbox{Apricot!50}{third best}.}
 \label{tab:euroc_ate}
\renewcommand\arraystretch{1.2} 
\setlength{\tabcolsep}{1.0mm}{ 
\resizebox{\textwidth}{!}{    
\begin{tabular}{c c cccccccccccc}\toprule
\multicolumn{2}{c}{Method} &MH01	&MH02	&MH03	&MH04	&MH05	&V101	&V102	&V103	&V201	&V202	&V203 &Avg \\ \midrule
\multirow{4}{*}{\rotatebox{90}{Stereo}} 
&ORB-SLAM2\cite{mur2017orb}  
&0.0350&0.0180&0.0280&0.1190&0.0600&0.0350&0.0200&0.0480&0.0370&0.0350&-&- \\
&ORB-SLAM3\cite{campos2021orb}    
&0.0290&0.0190&0.0240&0.0850&0.0520&0.0350&0.0250&0.0610&0.0410&0.0280&0.5210&0.0840\\
&DROID-SLAM\cite{teed2021droid}
&\cellcolor{Apricot!50}{0.0150}&\cellcolor{SpringGreen!50}{0.0130}&0.0350&0.0480&\cellcolor{SpringGreen!50}{0.0400}&0.0370&\cellcolor{Apricot!50}{0.0110}&\cellcolor{Yellow!50}{0.0200}&\cellcolor{Apricot!50}{0.0180}&0.0150&\cellcolor{Yellow!50}{0.0170}&\cellcolor{Apricot!50}{0.0240} \\
&GO-SLAM\cite{zhang2023go}
&0.0160&\cellcolor{Apricot!50}{0.0140}&\cellcolor{Apricot!50}{0.0230}&\cellcolor{Yellow!50}{0.0450}&0.0450&0.0370&\cellcolor{Apricot!50}{0.0110}&0.0230&\cellcolor{SpringGreen!50}{0.0160}&\cellcolor{Yellow!50}{0.0100}&0.0220&\cellcolor{Apricot!50}{0.0240} \\

\midrule
\multirow{5}{*}{\rotatebox{90}{Mono.}}  
&ORB-SLAM\cite{mur2015orb}  
&0.0710&0.0670&0.0710&0.0820&0.0600&\cellcolor{SpringGreen!50}{0.0150}&\cellcolor{SpringGreen!50}{0.0020}&-&0.0210&0.0180&-&- \\
&ORB-SLAM3\cite{campos2021orb}    
&0.0160& 0.0270& 0.0280& 0.1380& 0.0720&\cellcolor{Yellow!50}{0.0330}& 0.0150& 0.0330& 0.0230& 0.0290&-&-        \\
&DROID-SLAM\cite{teed2021droid}        
&\cellcolor{Yellow!50}{0.0130}&\cellcolor{Apricot!50}{0.0140}&\cellcolor{Yellow!50}{0.0220}&\cellcolor{SpringGreen!50}{0.0430}& \cellcolor{Apricot!50}{0.0430}& 0.0370& 0.0120&\cellcolor{Yellow!50}{0.0200}&\cellcolor{Yellow!50}{0.0170}& 0.0130&\cellcolor{SpringGreen!50}{0.0140}&\cellcolor{SpringGreen!50}{0.0220}        \\ 
&GO-SLAM\cite{zhang2023go}        
&0.0172&0.0186&3.3548&5.2975&4.9379&0.7667&0.0123&1.3245&0.0249&\cellcolor{Apricot!50}{0.0110}&\cellcolor{Apricot!50}{0.0179}&1.4348       \\ 
&Ours        
&\cellcolor{SpringGreen!50}{0.0104}&\cellcolor{Yellow!50}{0.0131}&\cellcolor{SpringGreen!50}{0.0201}&\cellcolor{Apricot!50}{0.0458}&\cellcolor{Yellow!50}{0.0413}&\cellcolor{Apricot!50}{0.0341}&\cellcolor{Yellow!50}{0.0082}&\cellcolor{Apricot!50}{0.0153}&0.0198&\cellcolor{SpringGreen!50}{0.0091}&0.0262&\cellcolor{Yellow!50}{0.0221}       \\ \bottomrule
\end{tabular}}}
\end{center}
\end{table*}
\begin{table*}[htb]
\begin{center}
\small
 \caption{
 \textbf{Impact of Global Optimization.}
 We report the Average ATE RMSE[m] on Different Datasets with/without performing Global Optimization.}
 \label{tab:ablation_ate}
\renewcommand\arraystretch{1.2} 
\setlength{\tabcolsep}{2.0mm}{ 
\resizebox{7.5cm}{!}{    
\begin{tabular}{c c ccccccccc}\toprule
&
&Replica\cite{straub2019replica}& ScanNet\cite{dai2017scannet}&TUM RGB-D\cite{sturm2012benchmark}   & EuRoC\cite{burri2016euroc}&  \\ \midrule
&w/o GO   
&0.0053&0.1427&0.1158&0.0967   \\
&w/ GO  
&\textbf{0.0032}&\textbf{0.0746}&\textbf{0.0537}&\textbf{0.0221} \\
\bottomrule
\end{tabular}}}
\end{center}
\end{table*}

\subsection{Quantitative Evaluation}
\subsubsection{Replica\cite{straub2019replica}.}
\cref{tab:replica_ate} and \cref{tab:replica_render} respectively show the trajectory accuracy and rendering quality results on the Replica\cite{straub2019replica} dataset.
As a deep-learning-based SLAM method that includes global optimization and loop closure, our method remains comparable with GO-SLAM\cite{zhang2023go} in terms of trajectory accuracy.
Furthermore, though based on monocular input, our method shows competitive performance with RGBD-based Gaussian SLAM\cite{keetha2023splatam} and outperforms previous NeRF-based methods\cite{yang2022vox,zhu2022nice,xu2022point}.
Note Gaussian Splatting SLAM\cite{matsuki2023gaussian} also supports monocular input but no monocular results are provided in its paper, we hope to make a comparison with it upon its open-source release.

As for rendering quality results, we use estimated poses to get rendered images of post-processed maps and compute the rendering metrics with the ground truth image.
The results show that our RGB-based method outperforms early RGB-D based NeRF SLAM methods\cite{zhu2022nice,yang2022vox} in almost all the metrics, and only slightly lower in PSNR than SOTA RGB-D based methods\cite{keetha2023splatam,sandstrom2023point}.

\subsubsection{ScanNet\cite{dai2017scannet}. }
We perform comparisons on several representative sequences from the ScanNet dataset, the results are shown in \cref{tab:scannet_ate}.
The results show that the difference between our method and SOTA deep-learning-based methods\cite{teed2021droid,zhang2023go} is less than 1cm,
we assume this is due to the presence of global optimization in all three methods.
Apart from that, our method has more than 40\% improvement compared to the SOTA Gaussian method\cite{keetha2023splatam} even though it has additional depth inputs.

\subsubsection{TUM RGB-D\cite{sturm2012benchmark}}
Apart from the methods above, we perform additional comparisons with other traditional RGBD SLAM methods\cite{whelan2015elasticfusion,mur2017orb} on TUM RGB-D benchmark, results are shown in \cref{tab:tum_ate}.
Unprovided results of DROID-SLAM\cite{teed2021droid} and GO-SLAM\cite{zhang2023go} are obtained by the aforementioned approach.

The results of our method are inferior to GO-SLAM\cite{zhang2023go} and we suppose that it may caused by the degradation of the loop closure module in over-dense indoor scenes.
Nevertheless, our method still maintains comparable performance with the SOTA RGB-D Gaussian SLAM methods\cite{keetha2023splatam}.
When compared with \cite{matsuki2023gaussian}, which is also an RGB-based method, our method has at least 1-2 cm of lift in all the scenarios it provided.

\subsubsection{EuRoC\cite{burri2016euroc}}
In EuRoC\cite{burri2016euroc} dataset, since the depth information is unavailable, we compare our method with both monocular and stereo methods\cite{mur2017orb,mur2015orb,campos2021orb,zhang2023go,teed2021droid}, the results are shown in \cref{tab:euroc_ate}.
It is noteworthy that in some scenarios where the ATE remained extremely large no matter how many trials we performed, we assume that it may fail in these scenarios.
The results show that we achieve better performance than current monocular methods in most of the scenarios, even outperforming some of the results obtained in stereo settings.
Note that the average accuracy of our method and SOTA method\cite{teed2021droid} is about the same (less than 1mm difference).

\subsection{Qualitative Results}

\subsubsection{Rendering Quality.}
\cref{fig:render_result} shows the results of our method in terms of rendering quality.
Thanks to the introduction of 3D Gaussian representation, our method exhibits excellent performance in rendering quality that exceeds previous NeRF-based methods\cite{xu2022point,zhu2022nice,zhang2023go}.
Note that \cite{keetha2023splatam} is also based on 3D Gaussian representation, but since it has additional dense depth information, we do not make a comparison with it here.
The method mainly compared here is GO-SLAM\cite{zhang2023go} cause it is also based on monocular input and has a loop closure module, our method can achieve better rendering results while maintaining comparable tracking accuracy.

\subsubsection{Text-to-trajectory Query.}
In \cref{fig:query_result}, we demonstrate the examples of text-to-trajectory querying.
The results indicate that given a text prompt, our system can calculate the similarity score along the trajectory and return the most relevant keyframe.
Such advanced functionality facilitates our system for more applications that require a high-level understanding of the environment.

\subsection{Ablation Study}
In order to study the effectiveness of global optimization based on Back-End Graph in improving trajectory accuracy, we perform ablation experiments on multiple datasets.
The result in \cref{tab:ablation_ate} shows that the global optimization brings more than 50\% accuracy improvement on several datasets.

\section{Conclusion}
In this paper, we present MG-SLAM, a monocular Gaussian SLAM that uses 3D Gaussian as a global map representation, and introduce a language-extended loop closure module to correct drift errors.
We demonstrate its effectiveness in drift-corrected tracking and photo-realistic mapping.
Further, we also show its potential for achieving a high-level understanding of the scene while performing loop closure, which reveals the possibility of more advanced applications such as embodied artificial intelligence in the future.

%
%
\bibliographystyle{splncs04}
\bibliography{main}
\end{document}